\documentclass[10pt,twocolumn,letterpaper]{article}

\usepackage{wacv}
\usepackage{times}
\usepackage{epsfig}
\usepackage{graphicx}
\usepackage{amsmath}
\usepackage{amssymb}
\usepackage{multirow}
\usepackage{makecell}
\usepackage[table]{xcolor}
\usepackage{fancyhdr}
\usepackage{arydshln}
\usepackage{animate}
\usepackage{array}
\usepackage{xr}

\usepackage[ruled,vlined]{algorithm2e}
\usepackage{times}
\usepackage{epsfig}
\usepackage{graphicx}
\usepackage{amsmath}
\usepackage{amssymb}
\usepackage[utf8]{inputenc} 
\usepackage[T1]{fontenc}    
\usepackage{url}            
\usepackage{booktabs}       
\usepackage{amsfonts}       
\usepackage{nicefrac}       
\usepackage{microtype}      
\usepackage{xcolor}         
\usepackage{microtype}      
\usepackage{graphicx}
\usepackage{amsmath}
\usepackage{mathtools}
\usepackage{makecell}
\usepackage{amsthm}
\usepackage{enumitem}
\usepackage{subcaption}
\usepackage{booktabs}
\usepackage{ragged2e}
\usepackage{arydshln}
\usepackage[hang,flushmargin]{footmisc}
\usepackage{multirow}
\usepackage{multicol}
\usepackage{xr-hyper}
\usepackage[pagebackref=true,breaklinks=true,colorlinks,bookmarks=false]{hyperref}

\def\wacvPaperID{878} 

\wacvfinalcopy 

\ifwacvfinal
\def\assignedStartPage{9876} 
\fi

\ifwacvfinal
\usepackage[breaklinks=true,bookmarks=false]{hyperref}
\else
\usepackage[pagebackref=true,breaklinks=true,colorlinks,bookmarks=false]{hyperref}
\fi

\ifwacvfinal
\else
\fi

\begin{document}
\setlength{\abovedisplayskip}{3pt}
\setlength{\belowdisplayskip}{3pt}
\title{Perceptual Consistency in Video Segmentation}

\author{Yizhe Zhang \thanks{Work done at Qualcomm AI Research.} \thanks{Nanjing University of Science and Technology, Nanjing, China} \\\small{yizhe.zhang.cs@gmail.com}
\and
Shubhankar Borse \thanks{Qualcomm AI Research, an initiative of Qualcomm Technologies, Inc.} \thanks{Equal contribution.}\\\small{sborse@qti.qualcomm.com}
\and
Hong Cai \footnotemark[3] \footnotemark[4]\\\small{hongcai@qti.qualcomm.com}
\and
Ying Wang \footnotemark[1] \\\small{ yingwang0022@gmail.com}
\and
Ning Bi  \thanks{Qualcomm Technologies, Inc.}\\\small{nbi@qti.qualcomm.com}
\and
Xiaoyun Jiang  \footnotemark[5]\\\small{ xjiang@qti.qualcomm.com}
\and
Fatih Porikli \footnotemark[3]\\\small{fporikli@qti.qualcomm.com}}
\maketitle

\begin{abstract}
In this paper, we present a novel \textbf{perceptual consistency} perspective on video semantic segmentation, which can capture both temporal consistency and pixel-wise correctness. Given two nearby video frames, perceptual consistency measures how much the segmentation decisions agree with the pixel correspondences obtained via matching general perceptual features. More specifically, for each pixel in one frame, we find the most perceptually correlated pixel in the other frame. Our intuition is that such a pair of pixels are highly likely to belong to the same class. Next, we assess how much the segmentation agrees with such perceptual correspondences, based on which we derive the perceptual consistency of the segmentation maps across these two frames. Utilizing perceptual consistency, we can evaluate the temporal consistency of video segmentation by measuring the perceptual consistency over consecutive pairs of segmentation maps in a video. Furthermore, given a sparsely labeled test video, perceptual consistency can be utilized to aid with predicting the pixel-wise correctness of the segmentation on an unlabeled frame. More specifically, by measuring the perceptual consistency between the predicted segmentation and the available ground truth on a nearby frame and combining it with the segmentation confidence, we can accurately assess the classification correctness on each pixel. Our experiments show that the proposed perceptual consistency can more accurately evaluate the temporal consistency of video segmentation as compared to flow-based measures. Furthermore, it can help more confidently predict segmentation accuracy on unlabeled test frames, as compared to using classification confidence alone. Finally, our proposed measure can be used as a regularizer during the training of segmentation models, which leads to more temporally consistent video segmentation while maintaining accuracy.
\end{abstract}

\pagenumbering{gobble}
\section{Introduction}
High-quality video semantic segmentation is a critical task for a large variety of downstream applications, such as video processing, AR/VR, robotics, and autonomous driving. To be deployed for practical use, the segmentation model needs to be not only pixel-wise accurate, but also temporally consistent when applied to videos. In order to facilitate the development of such high-quality models, it is necessary to be able to assess both pixel-wise accuracy and temporal consistency accurately.

\begin{figure*}[t]
    \centering
    \includegraphics[width=0.99\linewidth]{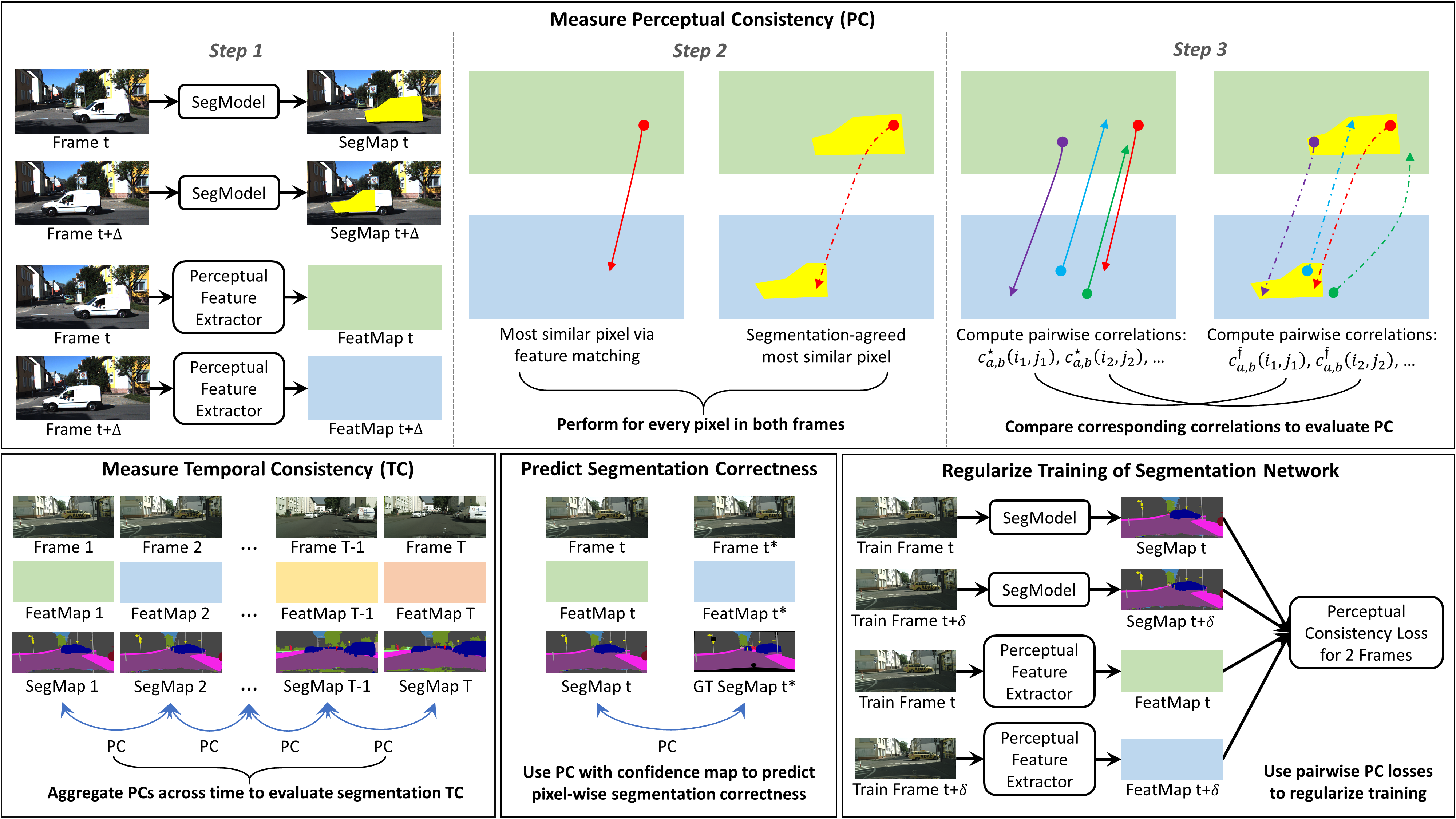}
    \caption{\small \textbf{Top:} Overview of measuring perceptual consistency of segmentation decisions on two nearby frames. 
    \textbf{Bottom:} Use cases of perceptual consistency, including measuring temporal consistency of video semantic segmentation, predicting pixel-wise segmentation correctness on unlabeled test video frames, and regularizing the training of a segmentation network. 
    }
\label{fig: overview}
\end{figure*}

While segmentation accuracy can be easily evaluated given the ground-truth labels on the test data, assessing the temporal consistency is not straightforward. In earlier work, Perazzi et al.~\cite{perazzi2016benchmark} measure the temporal consistency of object segmentation by computing the distance between the segmentation masks across two consecutive frames. This approach, however, does not factor in the object movements and changing occlusions. Most of the recent works~\cite{kundu2016feature, nilsson2018semantic, varghese2020unsupervised} utilize motion-based pixel correspondence between two consecutive frames (i.e., optical flow~\cite{ilg2017flownet}), to measure temporal consistency. More specifically, given two consecutive video frames, the segmentation of one frame is warped to the other based on the estimated flow, and the warped and actual segmentation maps are then compared to measure the segmentation consistency between these two frames. However, it is challenging to generate highly accurate and generalizable optical flow estimations. Furthermore, such exact pixel correspondence is susceptible to occlusions and objects moving out of the frame. 

When ground-truth annotations are not available during test/deployment, we can no longer use the standard accuracy metrics (e.g., mIoU, pixel-wise accuracy). For instance, in popular video semantic segmentation datasets such as Cityscapes~\cite{cordts2016cityscapes} and CamVid~\cite{brostow2008segmentation}, the test video frames are only sparsely labeled due to the high annotation costs. When ground-truth labels are not available on a test frame, one can use the pixel-wise classification confidence scores to predict the uncertainty or correctness of the segmentation network. However, this only captures the network's own output on the unlabeled frame and does not utilize the sparsely labeled frames.

In this work, we present a new perspective on the quality of video semantic segmentation models, i.e., \textbf{perceptual consistency}. Our intuition is that given two nearby video frames, which share similar visual contents, pixels that are highly perceptually similar are highly likely to belong to the same class and thus, should receive the same label from the segmentation model (SegModel). In order to concretize this, given two nearby video frames and the predicted segmentation maps (SegMaps) on them, we assess how much the segmentation agrees with the cross-frame pixel correspondence established on the two frames' perceptual feature maps (FeatMaps). More specifically, for each pixel in one frame, we first find the most correlated pixel from the other frame by matching perceptual features. We consider these two pixels are expected to belong to the same class. We next find the most correlated pixel that is also agreed by the segmentation maps. If the segmentation agrees with the perceptual correspondence, the correlations found via unconstrained feature matching and segmentation-agreed feature matching will be equal; otherwise, the segmentation-constrained one will be smaller. As such, we can use the ratio between these two correlations to quantify the (pixel-wise) agreement between the segmentation decisions and the perceptual correspondence. This ratio can then be aggregated over the pixels to measure the perceptual consistency between the segmentation maps on the two frames. This procedure is illustrated in Fig.~\ref{fig: overview} (top).

We can see that, unlike optical flow, our measure does not seek exact pixel correspondence across two images, which requires each pair of corresponding pixels to associate with the same object point; it instead finds pairs of maximally correlated pixels. This relaxation makes our measure robust to cases where exact correspondence does not exist, e.g., due to occlusions.

By measuring the perceptual consistency between segmentation maps on consecutive pairs of frames, we can naturally capture the temporal consistency of video segmentation. In addition, in the case of a sparsely labeled test video, we can utilize perceptual consistency (between an unlabeled frame and a nearby labeled frame), in addition to the segmentation confidence, to accurately predict the pixel-wise segmentation correctness on an unlabeled test frame. This allows the prediction to leverage information beyond a single frame. Furthermore, we can integrate perceptual consistency as an additional regularization in training (which does not incur extra computation during inference). This enables the trained video segmentation network to generate temporally more consistent results while maintaining accuracy. These use cases of perceptual consistency are illustrated in Fig.~\ref{fig: overview} (bottom).

We summarize our main contributions as follows:

\begin{itemize}
\item We propose a novel perceptual consistency measure that evaluates how much the segmented predictions over two similar images agree with the pixel correspondence found via matching general perceptual features.
\item Our perceptual consistency measure, when applied to segmented predictions on consecutive pairs of frames in a video, can more accurately evaluate the temporal consistency, as compared to the existing flow-based measure. 
\item Given a non-training image that does not have the ground-truth annotation, our measure can be used to accurately predict the pixel-wise correctness of the segmentation on this image by cross-referencing the available ground-truth of a similar image.
\item Furthermore, our proposed measure can be used as a regularizer during training, which leads to more temporally consistent segmentation models while maintaining accuracy; this property holds even when compared to models trained with optical flow.
\end{itemize}

\section{Related Work}\label{sec:related work}

\noindent \textbf{Video Semantic Segmentation:}
Video segmentation has been a major goal in computer vision research. In addition to improving the accuracy of semantic predictions~\cite{ding2020every, tao2020hierarchical, tian2020triple}, previous works have studied various other aspects that are more specific to the video setting, e.g., efficient processing~\cite{hu2020temporally, li2018low, nekrasov2020architecture}, label propagation~\cite{oh2019video, wang2019ranet}, and temporal consistency~\cite{hur2016joint, liu2020efficient, nilsson2018semantic}. We refer readers to recent surveys~\cite{minaee2020image, yao2019video} for a more comprehensive discussion.

\noindent \textbf{Temporal Consistency:}
Since accuracy measures (e.g., mIoU) cannot properly capture temporal consistency, researchers have proposed various ways to measure segmentation temporal consistency. For instance, \cite{perazzi2016benchmark} computes the distance between the segmentation masks from two consecutive frames. This, however, does not factor in motion. As such, researchers incorporate motion estimation (e.g., optical flow) when measuring temporal consistency~\cite{kundu2016feature, nilsson2018semantic, liu2020efficient, varghese2020unsupervised}.
However, estimating accurate flow on real-world data can be very challenging, and in many cases, more error-prone and time-consuming than the segmentation task itself. 

In order to improve temporal consistency, previous works have designed various models to take in additional information, e.g., optical flow~\cite{hur2016joint, nilsson2018semantic}, 3D structure~\cite{floros2012joint, kundu2014joint}, and temporal correlation~\cite{hu2020temporally, kundu2016feature, rebol2020frame, sibechi2019exploiting}. However, these methods require multi-frame information to segment each frame, incurring extra computation overhead. Recently, Liu et al.~\cite{liu2020efficient} have proposed using optical flow only during training and performing per-frame inference. However, their method is still constrained by the accuracy of the estimated flow.

\noindent \textbf{Predicting Segmentation Correctness:} Researchers have studied various ways to capture the output uncertainty/correctness of a neural network for classification tasks, e.g., maximum softmax probability (i.e., confidence score)~\cite{ hendrycks17baseline, guo2017calibration}, dropout/sampling~\cite{kendall2017uncertainties}, using ensembles~\cite{lakshminarayanan2017simple}, stochastic variational Bayesian inference~\cite{ovadia2019can}. However, during test time, these methods only utilize the information from the single unlabeled input image and can suffer from domain shift~\cite{ovadia2019can}. Moreover, except for the case of using classification confidence, the existing methods require specific training schemes and/or network modules. As such, they cannot be applied to any given trained model.


\section{Perceptual Consistency}\label{sec:method}
Perceptual consistency is our novel take on assessing the quality of video semantic segmentation. Given two nearby video frames, perceptual consistency measures how much the segmentation agrees with the cross-frame perceptual pixel correspondence. Perceptual consistency can be applied to the segmentation maps on consecutive pairs of frames in a video, in order to evaluate the segmentation temporal consistency. Furthermore, given a sparsely labeled test video, perceptual consistency can facilitate the prediction of pixel-wise correctness of the segmentation on a frame that does not have ground-truth annotation. Finally, we can utilize perceptual consistency to regularize the training of a segmentation network.

\subsection{Measuring Perceptual Consistency} \label{sec:measure pc}

Consider two nearby video frames which share similar visual contents, $x_a$ and $x_b$, and their respective segmentation maps (e.g., predicted by a neural network), $y_a$ and $y_b$. Let $f_a$ and $f_b$ denote the general perceptual feature maps for $x_a$ and $x_b$, which can be obtained from an image representation network trained on a large, general dataset (e.g., a ResNet~\cite{he2016deep} trained on ImageNet~\cite{deng2009imagenet}). 

\noindent \textbf{Perceptual Correspondence:} 
For every pixel $(i,j)$ in $x_a$, we find its perceptually corresponding pixel in $x_b$ by solving the following maximization:
\begin{equation}\label{eq:unconstrained}
    c_{a,b}^\star(i,j;\,f_a,f_b) = \underset{i',j'}{\text{max}}\ \sigma(f_a(i,j),\,f_b(i',j')),
\end{equation}
where $\sigma$ computes the cosine similarity between two feature vectors and $c_{a,b}^\star(i,j;\,f_a,f_b)$ is the maximum correlation between pixel $(i,j)$ in $x_a$ and the pixels in $x_b$, given the features maps, $f_a$ and $f_b$. We denote the most correlated pixel in $x_b$ as $(i^\star,j^\star)$.

The pixel $(i^\star,\,j^\star)$ in $x_b$ is the most perceptually similar to pixel $(i,\,j)$ in $x_a$. Since $x_a$ and $x_b$ are neighboring frames and share highly overlapping visual contents, it is highly likely that these two pixels belong to the class. For instance, on MIT DriveSeg~\cite{mit_driveseg}, a pair of perceptually corresponding pixels from two consecutive frames have a probability of $0.944$ to be in the same class. 

\noindent \textbf{Segmentation-Agreed Correspondence:}
Next, we find the pixel correspondence that is also agreed by the segmentation maps. More specifically, for pixel $(i,j)$ in $x_a$, we solve for the most perceptually correlated pixel in $x_b$ with the additional constraint that the two pixels are labeled with the same class on the segmentation maps:

\begin{equation}\label{eq:constrained}
\begin{split}
&c_{a,b}^\dagger(i,j;\,f_a, f_b, y_a, y_b) = \\
 &\underset{i',j'}{\text{max}}\ \sigma(f_a(i,j),\,f_b(i',j')),\ \ \text{s.t.}\ y_a(i,j) = y_b(i',j'),
\end{split}
\end{equation}

where $c_{a,b}^\dagger(i,j;\,f_a,f_b,y_a,y_b)$ is the segmentation-agreed maximum correlation between pixel $(i,j)$ in $x_a$ and the pixels in $x_b$, given the features maps, $f_a$ and $f_b$, and the segmentation maps, $y_a$ and $y_b$. we denote the found pixel in $x_b$ as $({i}^\dagger,{j}^\dagger)$.

\noindent \textbf{Measuring Consistency:}
If $i^\star=i^\dagger$ and $j^\star=j^\dagger$, then the segmentation decisions completely agree with the perceptual correspondence that these two pixels should belong to the same class. As a result, we have $c_{a,b}^\star(i,\,j) = c_{a,b}^\dagger(i,\,j)$.\footnote{We omit the given parameters listed after the semicolon when the context is clear.} Otherwise, $c_{a,b}^\dagger(i,\,j)$ will be smaller and its value depends on how much the segmentation decisions align with the perceptual correspondence. More specifically, if $x_b(i^\dagger,j^\dagger)$ is still highly perceptually correlated to $x_a(i,j)$ (while not being the most correlated one), then $c_{a,b}^\dagger$ will still be close to $c_{a,b}^\star(i,\,j)$. This indicates that the segmentation decisions highly agree with the perceptual correspondence. On the other hand, if $x_b(i^\dagger,\,j^\dagger)$ is not perceptually similar to $x_a(i,\,j)$, then $c_{a,b}^\dagger(i,\,j)$ will be much smaller than $c_{a,b}^\star(i,\,j)$. In this case, the segmentation decisions disagree with the perceptual correspondence, and do not assign the same class to $x_a(i,j)$ and its perceptually similar pixels in $x_b$. As such, by comparing $c_{a,b}^\dagger(i,\,j)$ and $c_{a,b}^\star(i,\,j)$, we can capture the pixel-wise consistency between the segmentation decisions and the perceptual correspondence. We refer to such a consistency as the \textbf{perceptual consistency}.

To quantify the perceptual consistency of the segmentation decisions over the two frames, we compute the ratio of $c_{a,b}^\dagger(i,\,j)$ and $c_{a,b}^\star(i,\,j)$, and then aggregate it over the frames as follows:
\begin{equation}\label{eq:pc_pair}
\begin{split}
    &\rho(y_a, y_b) = \\
    &\frac{1}{H\times W} \text{min} \Bigl\{\sum_{i,j}\frac{c_{a,b}^\dagger(i,j) - \bar{c}_{a,b}^\dagger}{c_{a,b}^\star(i,\,j)-\bar{c}_{a,b}^\star},\ \sum_{i,j}\frac{ c_{b,a}^\dagger(i,j) - \bar{c}_{b,a}^\dagger}{c_{b,a}^\star(i,\,j)-\bar{c}_{b,a}^\star} \Bigl\},
    \end{split}
\end{equation}
where $\bar{c}_{a,b}^\dagger$, $\bar{c}_{b,a}^\star$, $\bar{c}_{a,b}^\dagger$, and $\bar{c}_{b,a}^\star$ are the respective frame-level averages, and $H$ and $W$ are the frame height and width. We take the minimum of the two frame-level perceptual consistencies to better capture the disagreements between segmentation decisions and perceptual correspondences.

\subsection{Perceptual Consistency for Measuring Video Segmentation Temporal Consistency}
In a video (of length $T$), two consecutive frames share highly overlapping visual contents. As such, the perceptual consistency of Eq.~\ref{eq:pc_pair} can be readily applied to the segmentation maps of each consecutive pair of video frames. By aggregating the perceptual consistency over all the consecutive frame pairs, we obtain a measure of the temporal consistency of the video segmentation:
\begin{equation}\label{eq:pc_video}
    \tilde{\rho}(y_1,y_2,...,y_T) = \frac{1}{T-1}\sum_{t=1}^{T-1} \rho(y_t,y_{t+1}).
\end{equation}

\noindent \textbf{Robustness of Perceptual Consistency:}
Perceptual consistency offers a more robust way to measure temporal consistency as compared to optical flow, which most existing works use. For instance, when there is occlusion, optical flow will fail to find correspondence and thus cannot properly measure consistency for the affected image regions. On the other hand, perceptual consistency does not require each pair of perceptually corresponding pixels to associate with the same object point. Instead, we look for a pair of pixels that are the most perceptually correlated, which can always be found.

In addition, flow estimation errors can cause underestimation of the segmentation temporal consistency. For instance, when the estimated flow incorrectly connects two pixels that do not belong to the same class, the flow-based measure will assess zero consistency for this pair even when the segmentation decisions are correct. On the other hand, while it can occasionally happen that two perceptually corresponding pixels from two consecutive frames (found via Eq.~\ref{eq:unconstrained}) do not belong to the same class (e.g., with a probability of 0.056 on MIT DriveSeg), the probability is very high that given a pixel in one frame, there is a same-class pixel among the highly perceptually correlated pixels in the other frame. For instance, on MIT DriveSeg, this probability is $0.989$ if we consider the top-10 most correlated pixels, and the top-1 and top-10 correlation values differ only by $0.026$. As such, the segmentation model can still achieve a reasonably high consistency score as long as it correctly labels the perceptually similar pixels.

\subsection{Perceptual Consistency for Predicting Pixel-Wise Segmentation Correctness}\label{sec:method_estimate_accuracy}
In practice, the ground-truth labels may not be fully available during test. In such cases, we can predict the pixel-wise segmentation correctness on an unlabeled test frame by jointly using the segmentation confidence map (model-internal image-level measure) and the perceptual consistency map between the predicted segmentation on the test frame and the available ground truth of a nearby frame (model-external temporal-level measure).

Consider a sequence of test video frames, $X = \{x_1,x_2,...,x_T\}$, and the predicted segmentation maps, $Y = \{y_1,y_2,...,y_T\}$. The ground-truth segmentation maps are only available for a small subset of the frames, $S = \{s_t \ | \ t \in \Omega_L \}$, where $\Omega_L$ is the set of time instances where ground truths are available. This setup captures the structure of common video semantic segmentation datasets, such as Cityscapes~\cite{cordts2016cityscapes} and CamVid~\cite{brostow2008segmentation}
 
Given an unlabeled test frame, $x_{t_U}$, we can evaluate the quality of its segmentation by using both the decision confidence and perceptual consistency. For each pixel $(i,j)$, the confidence is given by $z_{t_U}(i,j) = \underset{k\in \Omega_K}{\text{max}} \tilde{z}_{t_U}(i,j,k)$, where $\tilde{z}_{t_U}(i,j,k)$ denotes the classification score for each class $k \in \Omega_K$ (after softmax). We then measure the pixel-wise perceptual consistency of $y_{t_U}$ w.r.t. the closest available ground-truth segmentation, $s_{t_L}$, where $t_L = \text{min}_{\tau \in \Omega_L} \|\tau - {t_U}\|$, as follows:
\begin{equation}\label{eq:PC map correctness}
    \hat{\rho}_{t_U}(i,j) = \frac{c_{t_U,t_L}^\dagger(i,j;\; f_{t_U},f_{t_L}) - \bar{c}_{t_U,t_L}^\dagger}{c_{t_U,t_L}^\star(i,\,j;\;f_{t_U},f_{f_L},y_{t_U},s_{t_L})-\bar{c}_{t_U,t_L}^\star},
\end{equation}
where the perceptual consistency is computed only in one direction from $x_{t_U}$ to $x_{t_L}$. This is because we only need to assess the consistency of the pixel-wise classification decisions on the unlabeled test frame w.r.t. the ground truth in this case.  

The final pixel-wise segmentation correctness is predicted by
\begin{equation}\label{eq:accuracy prediction}
    \alpha_{t_U}(i,j) = z_{t_U}(i,j) + \hat{\rho}_{t_U}(i,j),
\end{equation}
where perceptual consistency term allows the prediction to cross-reference the available ground truth of a nearby frame, in addition to using the network's confidence on the input unlabeled frame.

\subsection{Perceptual Consistency for Regularizing Network Training}\label{sec:training_regularization}
We can utilize perceptual consistency (which is differentiable) as an additional regularization when training any video segmentation network. This allows the network to learn to generate more perceptually consistent segmentation results across consecutive frames, which can improve temporal consistency. This is done without any change to the network. As such, our method provides an improvement without incurring extra computational cost during inference.

Given $N$ training videos, the perceptual consistency loss is given as follows:
\begin{equation}\label{eq:pc_loss}
    \mathcal{L}_{PC}= \frac{1}{N} \sum_{v=1}^{N} \frac{1}{T_v\times T_w}\sum_{i=1}^{T_v} \sum_{\delta =1}^{T_w} 1-\rho(x^v_t,x^v_{t+\delta}),
\end{equation}
where $\rho(x^v_t,x^v_{t+\delta})$ is computed using Eq.~\ref{eq:pc_pair} for frames $x^v_t$ and $x^v_{t+\delta}$ from the $v$-th video, $T_v$ is the length of video $v$, and $T_w$ is a time window within which we compute the loss for all the frame pairs.

The perceptual consistency loss of Eq.~\ref{eq:pc_loss} is then combined with the standard multi-class cross-entropy loss for segmentation~\cite{murphy2012machine}. The total loss for training the network is then given by
\begin{equation}\label{eq:total_loss} 
\mathcal{L} = \mathcal{L}_\text{CE} + \lambda \mathcal{L}_\text{TC},
\end{equation}
where $\lambda$ balances the cross-entropy loss and the perceptual consistency loss.

\section{Experiments}\label{sec:exp}
In this section, we conduct extensive experiments to demonstrate the efficacy of our proposed perceptual consistency, in terms of measuring the temporal consistency of video semantic segmentation, aiding the prediction of pixel-wise segmentation correctness, and regularizing the training of a segmentation network to promote better temporal consistency. We further conduct ablation studies to analyze the choice of perceptual features.

\subsection{Experiment Setup}
\noindent \textbf{Datasets:} For evaluating temporal consistency measures, we use MIT DriveSeg~\cite{mit_driveseg} and DAVIS 2016~\cite{perazzi2016benchmark} where per-frame ground-truth segmentation is available, and KITTI Flow 2015~\cite{menze2015object} where ground-truth optical flow is available. DriveSeg contains a video sequence of 5,000 frames with 15 semantic classes. The first 4,500 frames are used for training and the remaining 500 frames for test. DAVIS contains a total of 50 video sequences, with 3455 annotated frames. It is designed for foreground object segmentation. To measure TC, we use the validation set, consisting of 20 videos. In each frame, the ground-truth segmentation is provided to separate the object of interest from the background. For evaluating the prediction of segmentation correctness, we use DriveSeg. To evaluate the usage of perceptual consistency as a training regularization, we use Cityscapes~\cite{cordts2016cityscapes} and CamVid~\cite{brostow2008segmentation}, which are standard video semantic segmentation benchmarks.

\noindent \textbf{Perceptual Features:} In our main experiments, we use a ResNet-18~\cite{he2016deep} trained on ImageNet~\cite{deng2009imagenet} to extract perceptual features. We also include analysis of other feature extraction networks in the ablation studies, such as ResNet-101~\cite{he2016deep} and WideResNet-50 (WRN-50)~\cite{Zagoruyko2016WRN}.

\noindent \textbf{Segmentation Networks:} We use state-of-the-art segmentation models in our experiments. For evaluating TC, on DriveSeg, we use HRNet-w48~\cite{wang2020deep} and  PSPNet~\cite{zhao2017pyramid} with ResNet-101 backbone that are trained on DriveSeg. On DAVIS, we use the trained models of STM~\cite{oh2019video} and RANet~\cite{wang2019ranet} from the official repositories. On KITTI Flow, we use two HRNet-w48 models, one trained on DriveSeg and the other trained on Cityscapes. For evaluating the prediction of segmentation correctness, we use the PSPNet trained on DriveSeg. For the training experiments on Cityscapes and CamVid, we use HRNet-w18~\cite{wang2020deep} and DeepLabV3+~\cite{chen2018encoder} with ResNet-101 backbone.

\noindent \textbf{Temporal Consistency Measure:} We consider the flow-based TC as the ground-truth TC when using ground-truth flows, which computes the mIoU between warped and actual segmentation maps across frames. For a practical baseline, we use the flow-based measure in~\cite{liu2020efficient} which uses a pretrained FlowNetV2~\cite{flownet2-pytorch} to estimate optical flow.

\subsection{Measuring Temporal Consistency}
To evaluate the ability of perceptual consistency to measure segmentation temporal consistency (TC), we conduct experiments on the videos from DriveSeg and DAVIS. We further evaluate on KITTI Flow where we utilize the ground-truth optical flows to compute ground-truth TC.

\noindent \textbf{Measuring TC for Ground-Truth Segmentation:} We first perform a verification experiment where we measure the temporal consistency of ground-truth segmentation, for which an accurate measure should return a TC value close to 1. As shown in Table~\ref{tab:measure-GT-sequences}, perceptual consistency assesses a close-to-1 TC for the ground-truth video segmentation. On the other hand, the flow-based measure assesses much lower TC values for the ground truth.

\begin{table*}[t!]
\centering
\small
\begin{tabular}[h]{ | c | c | c | }
\hline
\textbf{Method} & {MIT DriveSeg}   &{DAVIS} 	\\
\hline
Flow-Based (FlowNetV2) & 0.8682 $\pm$ 0.0633  &  0.9454 $\pm$ 0.0540 \\
\hline
Perceptual Consistency & 0.9847 $\pm$ 0.0132 &0.9985 $\pm$ 0.0014 \\
\hline
\end{tabular}
\caption{\small Measuring TC of ground-truth segmentation on DriveSeg and DAVIS. We report the average TC values across time, as well as the standard deviations.}
\label{tab:measure-GT-sequences}
\end{table*}

\noindent \textbf{Evaluation Based on Per-Frame Ground-Truth Segmentation:} 
For a video, given the predicted segmentation, $y_1,y_2,...,y_T$, and the ground-truth segmentation, $s_1,s_2,...,s_T$, we can construct a sequence of alternating predicted and GT segmentation maps (e.g., $y_1, s_2,y_3,s_4,...$), for which the ground-truth TC can be derived. In this sequence, for a pair, $y_t$ and $s_{t+1}$, we can use $s_t$ to approximate the warped version of $s_{t+1}$ at $t$ under the assumption that one-to-one pixel correspondence exists between the two frames. The mIoU between $y_t$ and $s_t$ then captures the ground-truth TC between $y_t$ and $s_{t+1}$. We evaluate the correlation between perceptual consistency and the ground-truth TC for such alternating segmentation sequences. As shown in Tables~\ref{tab:results_driveseg}~and~\ref{tab:results_davis}, our perceptual consistency correlates much better with the ground-truth TC, in terms of the Pearson, Spearman, and Kendall correlation coefficients, as compared to the flow-based measure.

\begin{table*}[t!]
\centering
\small
\begin{tabular}[h]{ | c | c | c | c | c | c| c |}
\hline
\multirow{2}{*}{\textbf{Method}} & \multicolumn{2}{c|}{{Pearson}}   &\multicolumn{2}{c|}{{Spearman}} & \multicolumn{2}{c|}{{Kendall}} 	\\
&HRNet-w48&PSPNet&HRNet-w48&PSPNet&HRNet-w48&PSPNet \\
\hline
Flow-Based (FlowNetV2) &  0.5768  & 0.9459  & 0.5568 & 0.8568  & 0.4212 & 0.7118 \\
\hline
Perceptual Consistency &  0.7432 & 0.9492 & 0.8170 & 0.8602 & 0.6599 & 0.7293 \\
\hline
\end{tabular}
\caption{\small Correlations between measured TC and ground-truth TC on DriveSeg. A higher correlation indicates a more accurately measured TC. P-values for all reported correlations are smaller than $0.0001$.}
\label{tab:results_driveseg}
\end{table*}

\begin{table*}[t!]
\centering
\small
\begin{tabular}[h]{| c | c | c | c | c | c| c | }
\hline
\multirow{2}{*}{\textbf{Method}} & \multicolumn{2}{c|}{{Pearson}}   &\multicolumn{2}{c|}{{Spearman}} & \multicolumn{2}{c|}{{Kendall}} 	\\
&STM &RANet & STM &RANet &STM & RANet \\
\hline
Flow-Based (FlowNetV2) & 0.6627  & 0.6984  &0.7030& 0.7131 &  0.5218 & 0.5263 \\
\hline
Perceptual Consistency &  0.8048  & 0.8551 & 0.7297 & 0.8304 & 0.5393 & 0.6366\\
\hline
\end{tabular}
\caption{\small Correlations between measured TC and ground-truth TC on DAVIS. A higher correlation indicates a more accurately measured TC. P-values for all reported correlations are smaller than $0.0001$.}
\label{tab:results_davis}
\end{table*}

\noindent \textbf{Evaluation Based on Ground-Truth Flow:} We can obtain the ground-truth TC on KITTI Flow, by using the available ground-truth flow to warp the segmentation map from one frame to another and compute the mIoU between the warped and actual maps for every consecutive pair of frames. 
As shown in Table~\ref{tab:results_kitti}, in most cases, our perceptual consistency correlates significantly better with the ground-truth TC as compared to the measure using estimation flows.

\begin{table*}[t!]
\centering
\small
\begin{tabular}[h]{ | c | c | c | c | c | c| c |}
\hline
\multirow{2}{*}{\textbf{Method}} & \multicolumn{2}{c|}{{Pearson}}   &\multicolumn{2}{c|}{{Spearman}} & \multicolumn{2}{c|}{{Kendall}} 	\\
&HRN (C) &HRN (D) & HRN (C) &HRN (D) &HRN (C) & HRN (D) \\
\hline
Flow-Based (FlowNetV2) &  0.7124 & 0.7857  & 0.7654 & 0.7605 & 0.6028 &  0.5970 \\
\hline
Perceptual Consistency &  0.6808 &  0.8566 & 0.8022 & 0.8833 & 0.6113 & 0.7183\\
\hline
\end{tabular}
\caption{\small  Correlations between measured TC and ground-truth TC computed from ground-truth flows on KITTI Flow 2015. A higher correlation indicates a more accurately measured TC. P-values for all reported correlations are smaller than $0.0001$. HRN (C) and HRN (D) denote the HRNet-w48 models trained on Cityscapes and DriveSeg, respectively.}
\label{tab:results_kitti}
\end{table*}

\begin{figure*}[t!]
    \centering
    \includegraphics[width=0.80\linewidth]{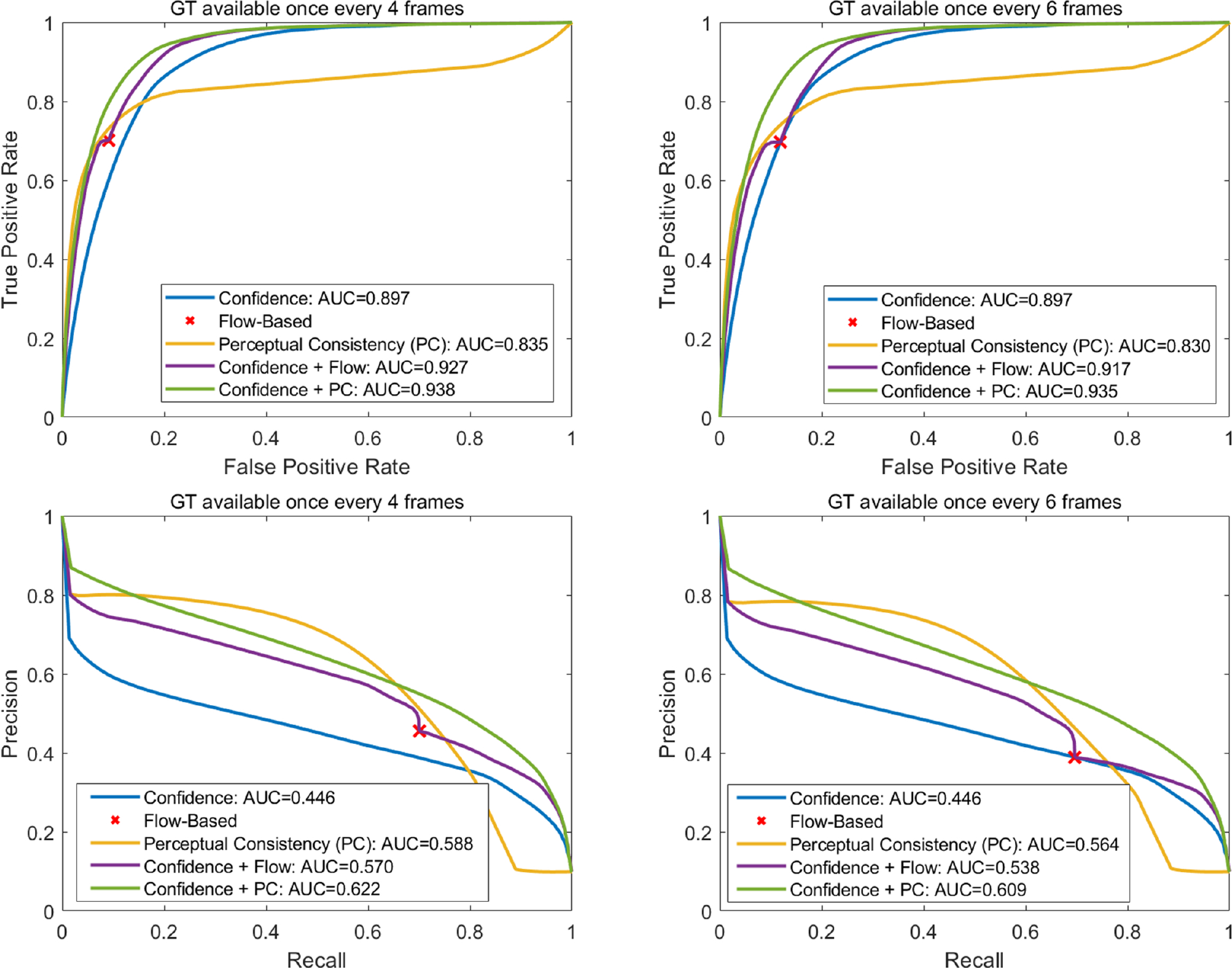}
    \caption{\small Receiver operating characteristic curves (top) and precision-recall curves (bottom) for predicting pixel-wise segmentation correctness on DriveSeg, for the cases where a ground-truth segmentation map is available every 4 frames (left) and every 6 frames (right). An incorrectly segmented pixel is considered as a positive sample and a correct one is a negative sample. The segmentation model is an PSPNet (with ResNet-101 backbone) trained on DriveSeg. 
    } 
\label{fig:roc-driveseg}
\end{figure*}

\subsection{Using Perceptual Consistency in Training}
\label{sec:TC_in_Training}
On Cityscapes and CamVid, we take a pretrained segmentation network (using only cross-entropy) and use the total loss of Eq.~\ref{eq:total_loss} (with $\lambda=0.5$) to finetune it for 20 epochs. We use stochastic gradient descent, and set the momentum to 0.9, weight decay to 0.0005, and learning rate to 0.0001. 

It can be seen in Table~\ref{tab:training} that by using our perceptual consistency regularization, we can improve the temporal consistency of the segmentation network while slightly improving or maintaining the accuracy. As compared to using a flow-based regularization~\cite{liu2020efficient}, perceptual consistency is more effective in promoting the model's temporal consistency. Note that we use the flow-based measure (with FlowNetV2) in~\cite{liu2020efficient} to evaluate temporal consistency, in order to compare with reported results in the literature (although the estimated flows can contain errors).

\subsection{Predicting Pixel-Wise Segmentation Correctness}

We use Eq.~\ref{eq:accuracy prediction} to predict the pixel-wise correctness of the segmentation, which fuses the classification confidence map and the perceptual consistency map of Eq.~\ref{eq:PC map correctness}. In Fig.~\ref{fig:roc-driveseg}, we show the Receiver Operating Characteristic (ROC) and precision-recall (PR) curves of our correctness prediction, for different sparsity levels of the available ground-truth segmentation maps. 
It can be seen that it is not sufficient to the confidence alone for the prediction (blue), since it only utilizes the information from a single unlabeled test image. Adding perceptual consistency considerably improves the prediction by exploiting the information from a nearby labeled frame (green). For instance, when a ground-truth segmentation map is available every 4 frames, adding perceptual consistency improves the prediction AUC from 0.897 to 0.938. This means that even if only 25\% of the video frames are annotated, our approach can still accurately predict the pixel-wise segmentation correctness for the unlabeled frames. It is noteworthy that confidence and perceptual consistency capture different characteristics of the correctness, as can be seen in the figure. As such, fusing them enhances the overall prediction.

Since optical flow also captures inter-frame information, we experiment with combining confidence and the flow-based measure to predict correctness (purple). This prediction is less accurate as compared to the case of adding perceptual consistency. Moreover, since the flow-based measure returns only a binary result for each pixel, it has only one point in the ROC plot. Such binary inputs introduce irregularity into the prediction, as shown by the less regular shape of the purple ROC curve.

\begin{table*}
\small
\centering
  \begin{tabular}{|c|c|c|c|c|}
    \hline
    \multirow{2}{*}{\textbf{Model and Training Scheme}} & 
      \multicolumn{2}{c}{Cityscapes} & 
      \multicolumn{2}{c|}{CamVid} \\
    & {mIoU} & Flow-Based TC  & mIoU & Flow-Based TC   \\
\hline
HRNet-w18~\cite{wang2020deep} & 0.762 & 0.691 & 0.732 &0.752\\ 
+ Flow-Based Loss~\cite{liu2020efficient} & 0.764 &0.696 &--   &-- \\ 
\textbf{+ Perceptual Consistency}& 0.764 &\textbf{0.712} &0.732  &\textbf{0.762} \\ 
\hline
DeepLabV3+ (ResNet-101) \cite{chen2018encoder}  & 0.762  & 0.710 &0.752 & 0.756 \\
\textbf{ + Perceptual Consistency} & 0.763 & \textbf{0.724} & 0.752 & \textbf{0.765}\\   
\hline
\end{tabular}
\caption{\small Training performance of utilizing perceptual consistency as an additional regularization, as compared to the cases of no regularization and using flow-based regularization~\cite{liu2020efficient}. In this evaluation, we use the flow-based measure (with FlowNetV2) in~\cite{liu2020efficient} to assess temporal consistency in order to compare with reported results in the literature (although the estimated flows can contain errors).}
\label{tab:training}
\end{table*}



\subsection{Additional Study}\label{sec:ablation}
In this part, we study the effect of using different perceptual feature extraction networks. We repeat the experiments of Table~\ref{tab:results_driveseg} and Table~\ref{tab:results_kitti} using different ImageNet-trained networks to extract feature maps for computing perceptual consistency. We use a DriveSeg-trained HRNet-w48 as the segmentation model. It can be seen in Table~\ref{tab:ablation_features} that when using different perceptual features, the TC measured by perceptual consistency has significantly higher correlations with the ground-truth TC, as compared to the flow-based measure.

\begin{table*}[t!]
\centering
\small
\begin{tabular}[h]{ | c | c | c | c | c | c | c | }
\hline
\multirow{2}{*}{\textbf{Method}} & \multicolumn{2}{c|}{{Pearson}}   &\multicolumn{2}{c|}{{Spearman}} & \multicolumn{2}{c|}{{Kendall}} 	\\
& DriveSeg & KITTI & DriveSeg & KITTI & DriveSeg & KITTI \\
\hline
Flow-Based (FlowNetV2)    &0.5768 &0.7857 &0.5568 &0.7605 &0.4212 &0.5970\\
\hline
Perceptual Consistency (ResNet-18)   &0.7432 &0.8566 &0.8170 &0.8833 &0.6599 &0.7183 \\
\hline
Perceptual Consistency (ResNet-101)  &0.7245 &0.8127 &0.8030 &0.8085 &0.6423 &0.6273 \\
\hline
Perceptual Consistency (WRN-50) &0.7448 &0.8504 &0.8188 &0.8583 &0.6608 &0.6896 \\
\hline
\end{tabular}
\caption{\small Correlations between measured TC and ground-truth TC. We show the performance of our perceptual consistency when using different feature extractors, such as ResNet-18, ResNet-101, and WRN-50. DriveSeg denotes the evaluation setting of Table~\ref{tab:results_driveseg} and KITTI denotes the evaluation setting of Table~\ref{tab:results_kitti}.}
\label{tab:ablation_features}
\end{table*}

\section{Discussion}

Above, we demonstrated our Perceptual Consistency (PC) is robust and accurate in measuring consistency and accuracy for video segmentation. Here, we discuss limitations and potential future work. PC requires finding dense matching for the perceptual features along the two directions, which at this moment is expected to be performed on a modern GPU/TPU for a decent throughput (e.g., under 100 ms per query). To demonstrate the robustness and generalization of PC, no special designed local window is used for the matching part in PC. However, if additional prior knowledge is available of the videos in hand, one may consider designing a way to dynamically change the search window for the matching step to save time and computation. To further improve the robustness of PC, one can use multiple representation networks to derive more robust and general perceptual features. A new self-supervised learning task might be possible for learning better feature representations for measuring perceptual consistency and beyond. In addition, it may be possible to cross-reference (with proper weighting) multiple GT-available frames to enhance prediction accuracy for predicting segmentation correctness.

\section{Conclusion}

This paper proposed a novel perceptual consistency measure to capture inconsistency and inaccuracy in video segmentation. Our framework draws its quality measure via cross-referencing segmentation among similar-looking images and builds a matching paradigm that can effectively and accurately detect and measure inconsistency and inaccuracy in segmentation. Combining network confidence measure (a model-internal image-level measure) and our perceptual consistency measure (a model-external temporal-level measure), we demonstrated that it is feasible to achieve high prediction accuracy in predicting pixel-wise segmentation consistency and correctness for video segmentation.   

{\small
\bibliographystyle{ieee_fullname}
\bibliography{egbib}
}

\clearpage
\section{Supplementary Materials}

\subsection{Predicting Segmentation Correctness across Datasets}\label{sec:predict_error_cross_dataset}

To further evaluate the efficacy of using our proposed perceptual consistency in predicting pixel-wise segmentation correctness, we conduct experiments with networks (HRNet-w18 and HRNet-w48) that are trained on Cityscapes and then tested on MIT DriveSeg. The network weights are obtained directly from the HRNet official repository.\footnote{The repository can be found at \url{https://github.com/HRNet/HRNet-Semantic-Segmentation/tree/pytorch-v1.1}.} 

\begin{figure}[h]
    \centering
    \includegraphics[width=0.99\linewidth]{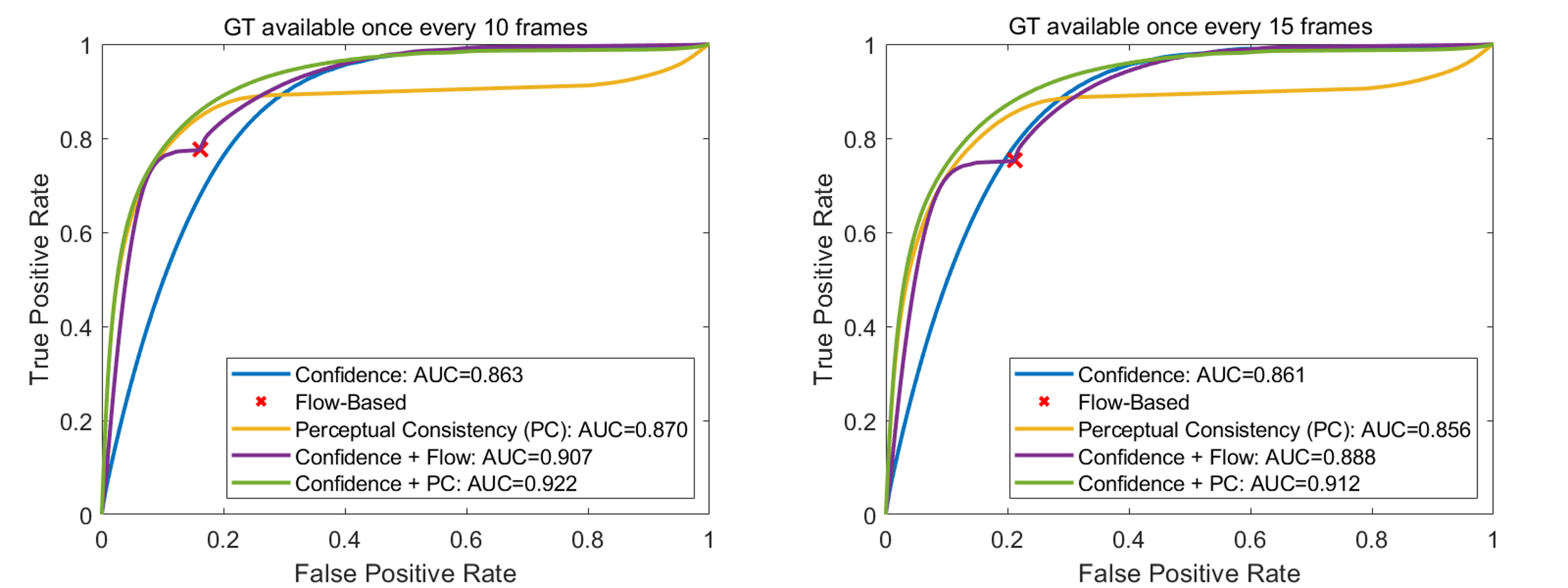}
    \caption{\small ROC curves for predicting pixel-wise segmentation correctness on DriveSeg, for the cases where a ground-truth segmentation map is available every 10 frames (left) and every 15 frames (right).  
    An incorrectly segmented pixel is considered as a positive sample and a correct one is a negative sample. The segmentation model is an HRNet-w18 trained on Cityscapes.
    } 
\label{fig:roc-cs-ds-hr18}
\end{figure}

\begin{figure}[h]
    \centering
    \includegraphics[width=0.99\linewidth]{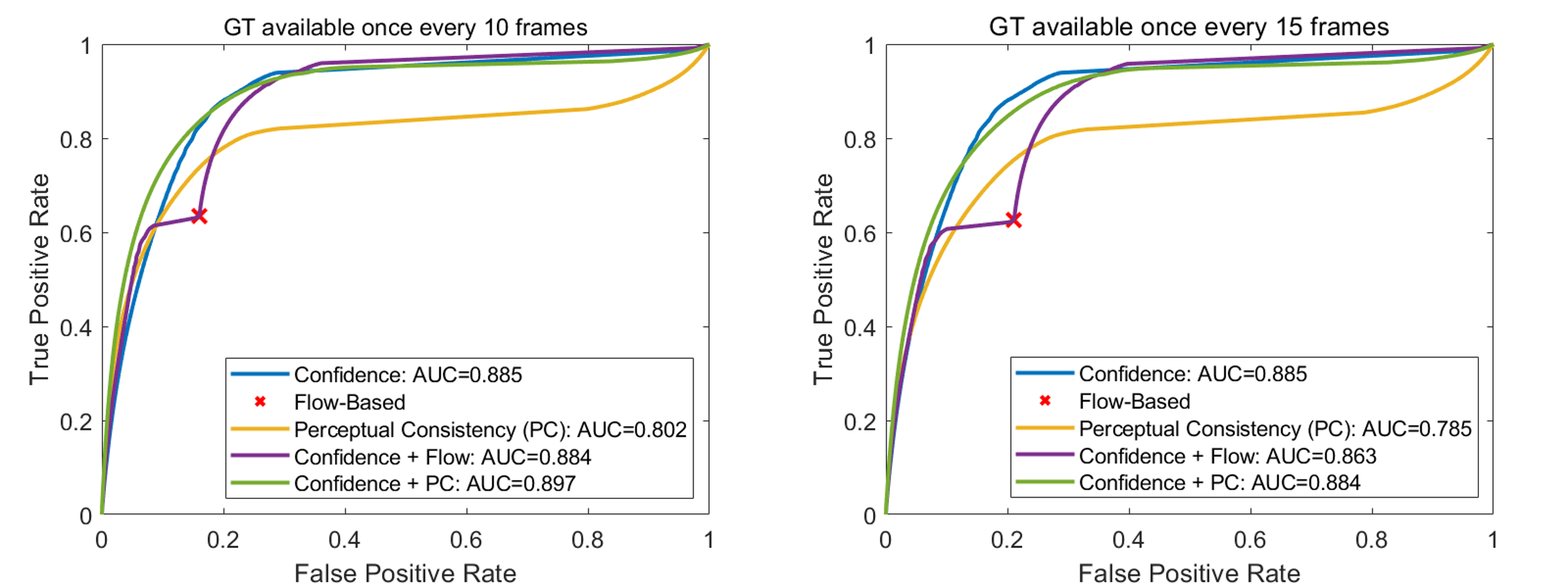}
    \caption{\small ROC curves for predicting pixel-wise segmentation correctness on DriveSeg, for the cases where a ground-truth segmentation map is available every 10 frames (left) and every 15 frames (right). An incorrectly segmented pixel is considered as a positive sample and a correct one is a negative sample. The segmentation model is an HRNet-w48 trained on Cityscapes.
    } 
\label{fig:roc-cs-ds-hr48}
\end{figure}

Fig.~\ref{fig:roc-cs-ds-hr18}~and~\ref{fig:roc-cs-ds-hr48} show the ROC curves for the cases of HRNet-w18 and HRNet-w48, respectively, with different sparsity levels of the available ground-truth segmentation maps. It can be seen that in all cases, by combining confidence and our perceptual consistency (green), we are able to improve the prediction accuracy of the segmentation correctness, as compared to using confidence alone (blue) and combining confidence and optical flow (purple).

\begin{figure}[h]
    \centering
    \includegraphics[width=0.99\linewidth]{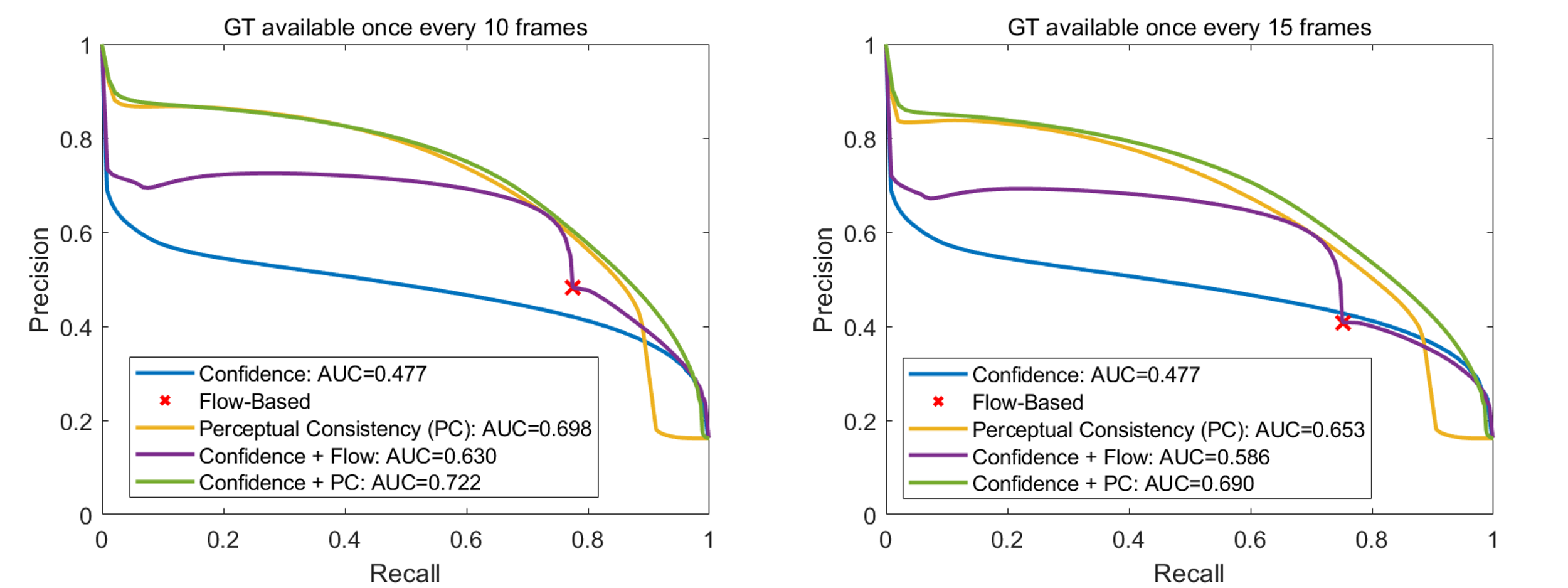}
    \caption{\small Precision-recall curves for predicting pixel-wise segmentation correctness on DriveSeg, for the cases where a ground-truth segmentation map is available every 10 frames (left) and every 15 frames (right). 
    An incorrectly segmented pixel is considered as a positive sample and a correct one is a negative sample. The segmentation model is an HRNet-w18 trained on Cityscapes. The AUC of a random classifier is 0.162.} 
\label{fig:pr-cs-ds-hr18}
\end{figure}

\begin{figure}[h]
    \centering
    \includegraphics[width=0.99\linewidth]{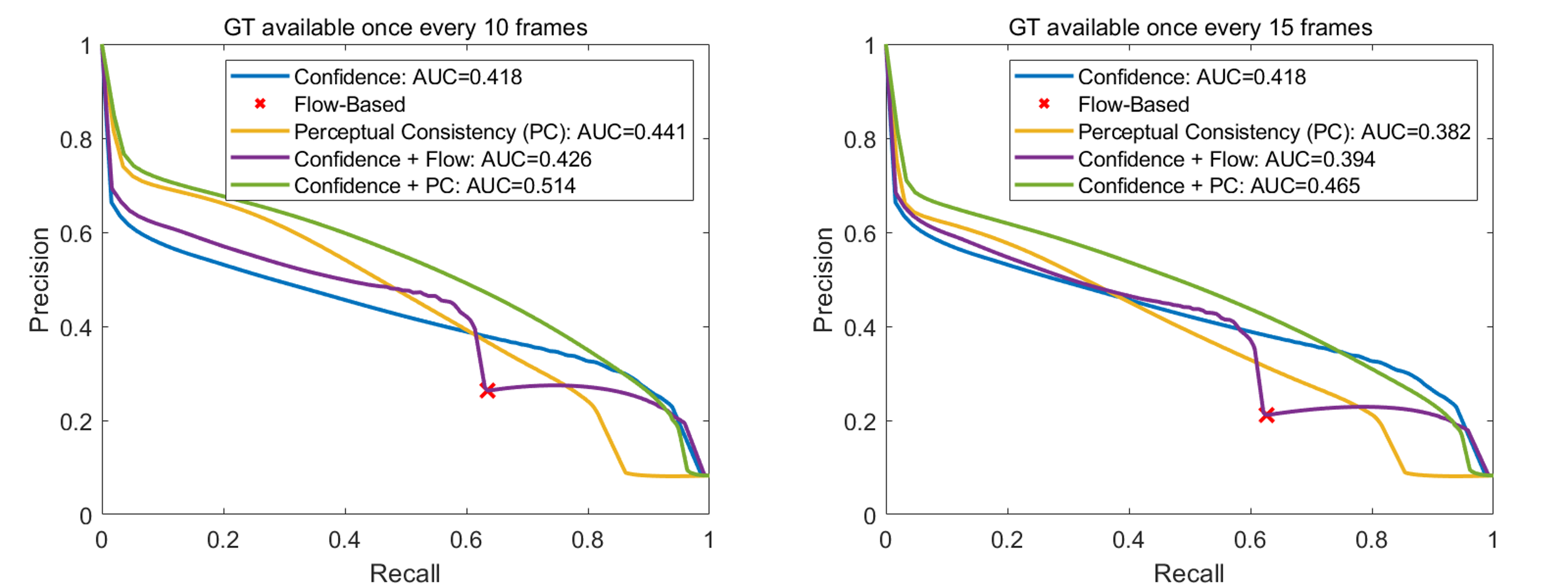}
    \caption{\small Precision-recall curves for predicting pixel-wise segmentation correctness on DriveSeg, for the cases where a ground-truth segmentation map is available every 10 frames (left) and every 15 frames (right). 
    An incorrectly segmented pixel is considered as a positive sample and a correct one is a negative sample. The segmentation model is an HRNet-w48 trained on Cityscapes. The AUC of a random classifier is 0.083.} 
\label{fig:pr-cs-ds-hr48}
\end{figure}

Fig.~\ref{fig:pr-cs-ds-hr18}~and~\ref{fig:pr-cs-ds-hr48} show the precision-recall curves. It can be seen that for the cases of both networks and for different sparsity levels of the ground truths, our proposed combination of confidence and perceptual consistency (green) provides significantly better prediction performance (in terms of AUC), as compared to using confidence alone (blue) and fusing confidence and optical flow (purple). Note that a random classifier for incorrect/correct segmentation will have precision-recall AUCs of 0.162 and 0.083 for the cases of HRNet-w18 and HRNet-w48, respectively.

\subsection{Visualizing Segmentation Error Prediction}
\begin{figure}[h]
    \centering
    \includegraphics[width=0.95\linewidth]{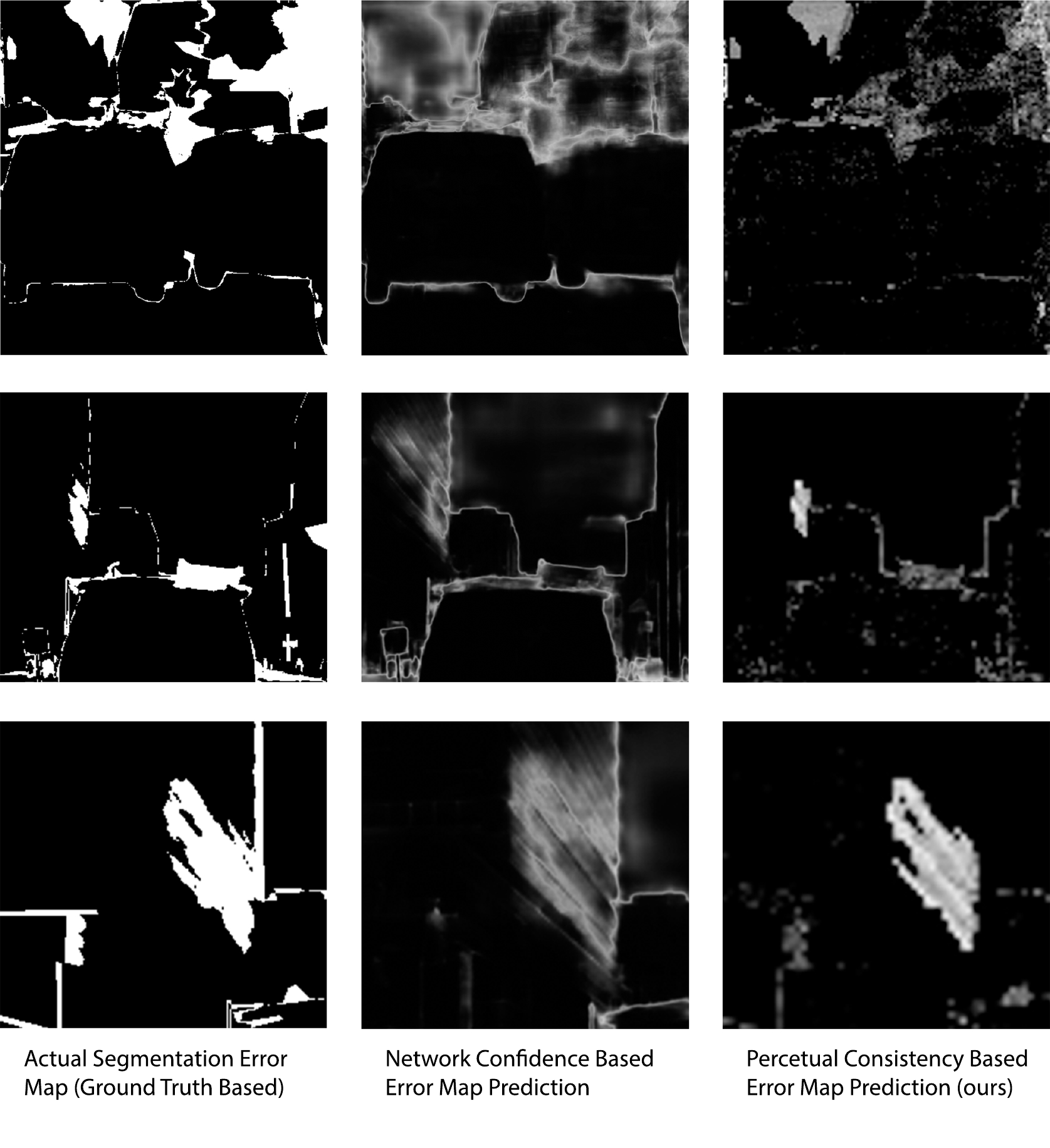}
    \caption{\small Visualization of ground-truth segmentation error map, confidence-based error prediction, and perceptual-consistency-based prediction. These results are based on the experiments of Sec.~\ref{sec:predict_error_cross_dataset} in this supplementary file (with the HRNet-w18 model).
    } 
\label{fig:vis_error_maps}
\end{figure}

In Fig.~\ref{fig:vis_error_maps}, we visually compare the true segmentation errors (first column), the confidence-based predicted error map (second column), as well as the predicted error map based on our perceptual consistency (third column). It can be seen that the confidence-based prediction and the perceptual-consistency-based prediction complement each other. This is because the confidence captures the model-internal image-level information while perceptual consistency captures the model-external temporal information. 


\end{document}